\documentclass[conference]{IEEEtran}
\IEEEoverridecommandlockouts
\pdfoutput=1
\usepackage{cite}
\usepackage{amsmath,amssymb,amsfonts}
\usepackage{algorithm}
\usepackage{algorithmic}
\usepackage{graphicx}
\usepackage{textcomp}
\usepackage{xcolor}
\usepackage{url}
\usepackage{comment}
\usepackage{booktabs}
\usepackage{multirow}
\usepackage{multicol}
\usepackage[font=small]{caption}

\def\BibTeX{{\rm B\kern-.05em{\sc i\kern-.025em b}\kern-.08em
    T\kern-.1667em\lower.7ex\hbox{E}\kern-.125emX}}
\begin{document}

\title{Visual Saliency-Guided Channel Pruning\\for Deep Visual Detectors in Autonomous Driving}


\author{
\IEEEauthorblockN{Jung Im Choi}
\IEEEauthorblockA{\textit{Department of Computer Science} \\
\textit{Bowling Green State University}\\
Bowling Green, USA \\
choij@bgsu.edu}
\and
\IEEEauthorblockN{Qing Tian}
\IEEEauthorblockA{\textit{Department of Computer Science} \\
\textit{Bowling Green State University}\\
Bowling Green, USA \\
qtian@bgsu.edu}
}

\newcommand{\etal}{\textit{et al}.}
\bibliographystyle{ieeetr}

\maketitle

\begin{abstract}
Deep neural network (DNN) pruning has become a de facto component for deploying on resource-constrained devices since it can reduce memory requirements and computation costs during inference. In particular, channel pruning gained more popularity due to its structured nature and direct savings on general hardware. However, most existing pruning approaches utilize importance measures that are not directly related to the task utility. Moreover, few in the literature focus on visual detection models. To fill these gaps, we propose a novel gradient-based saliency measure for visual detection and use it to guide our channel pruning. Experiments on the KITTI and COCO\_traffic datasets demonstrate our pruning method's efficacy and superiority over state-of-the-art competing approaches. It can even achieve better performance with fewer parameters than the original model. Our pruning also demonstrates great potential in handling small-scale objects.
\end{abstract}

\begin{IEEEkeywords}
Channel pruning, Visual saliency, Deep visual detection
\end{IEEEkeywords}

\section{Introduction}

Deep neural networks 
are powerful, but their large size and high computation requirements make it challenging to deploy them onto mobile/embedded devices (e.g., dash cams) or in scenarios where real-time inference is required (e.g., autonomous driving).
Network compression has been widely studied to reduce memory and computational costs. Popular network compression techniques include quantization~\cite{gupta2015deep}, knowledge distillation~\cite{Hinton2015}, and network pruning~\cite{Han2015,He2019}. Unlike other neural network compression approaches, pruning directly removes network components. They usually take in a pre-trained model, prune, and fine-tune to regain performance. The main differences between various pruning methods lie in pruning granularity and the importance measure.

Unstructured pruning directly removes individual parameters/connections. The resulting unstructured sparsity requires specialized software~\cite{Park2017} and hardware~\cite{Han2016b} to achieve real acceleration of the pruned model.
In contrast, structured pruning removes entire channels/filters/neurons~\cite{Li2017,Molchanov2017,Tian2017deep,He2019,Tian2021task}, leading to structured sparsity that can be directly utilized by general-purpose hardware. Such pruned networks can not only reduce the storage space but also speed up the inference. In this paper, we focus on structured channel pruning.

Different filter/channel pruning approaches usually differ in their importance measures.
Although existing methods have achieved promising results, most of them use ad-hoc human-defined or locally computed importance measures that are not directly related to the final task utility. Moreover, few focus on the visual detection task. In this paper, we propose a gradient-based saliency measure for visual detection and use it to guide the pruning. A channel with high gradient-based detection saliency contributes more to the final prediction of the visual detector. The main contributions of our work can be summarized as follows:
\begin{itemize}
    \item Unlike most existing pruning methods, we utilize the gradients of the detection utility with respect to channel features as our saliency measure to prune deep visual detectors in autonomous driving scenarios. 
    \item We incorporate ground truth bounding box information and their surrounding context information to reweigh the detection task gradients, which further improves the pruning performance. 
    \item Extensive experiments on the KITTI and COCO\_traffic datasets show that our pruning can outperform many state-of-the-art methods. Moreover, it can even improve the base model's performance with fewer parameters (e.g., for YOLOX-S on the KITTI dataset, it improves 1.8\% mAP over the unpruned model with 40.2\% and 29.2\% model size and FLOPs reductions, respectively). Our approach also demonstrates great potential in handling small-scale objects. 
\end{itemize}

\section{Related Work}
\label{related}

\subsection{Neural Network Pruning}

Based on the sparsity induced, pruning methods can be categorized into unstructured pruning and structured pruning.
\subsubsection{Unstructured Pruning}

Optimal Brain Damage~\cite{LeCun1989} and Optimal Brain Surgeon~\cite{Hassibi1992} are considered early works that prune weights individually. Although these approaches demonstrated that unnecessary weights could be removed from a trained network with little accuracy loss, they were not suitable to more recent deep neural networks due to intensive computational cost~\cite{Hassibi1992}. Ever since, various unstructured pruning methods have been proposed~\cite{Han2015,Srinivas2015,Chen2015,Zhang2018}. Such approaches usually set unimportant weights to zero and freeze them during re-training. Utilizing such unstructured sparsity requires specialized software~\cite{Park2017} and hardware support~\cite{Han2016b}.

\subsubsection{Structured Pruning}

Filter/channel pruning gained more popularity in recent years due to its direct savings on general hardware~\cite{Hu2016,Tian2017deep,He2019,Lin2020,Molchanov2017}. Our proposed method falls within this line of research. The key difference among them is the way to quantify the importance of a filter/channel. Several pioneering works in filter/channel pruning used simple metrics like the L1-norm~\cite{Li2017} / L2-norm~\cite{He2018} of filter weights to evaluate their importance. He~\etal~\cite{He2019} proposed to use geometric-median-based criterion to identify the most replaceable filters to be pruned. 

Another group of pruning studies focuses on features produced by filters~\cite{Hu2016,Lin2020,Sui2021}, rather than filters themselves. Hu~\etal~\cite{Hu2016} used the average percentage of zeros (APoZ) in the output feature maps. Lin~\etal~\cite{Lin2020} utilized the rank of the feature maps to prune filters generating low-rank feature maps. Sui~\etal~\cite{Sui2021} used channel independence by measuring the correlation among different feature maps to evaluate the importance of filters. However, these previously mentioned methods utilized locally computed importance measures that are not aware of the final task utility. 

In addition, while most of the existing channel pruning works have focused on the classification task~\cite{Li2017,Lin2020,Sui2021}, there have been relatively few works focused on the object detection task~\cite{Xie2020,Li2022}. Xie~\etal~\cite{Xie2020} presented a localization-aware auxiliary network to find out the channels with location-aware information for object detection and Li~\etal~\cite{Li2022} proposed a multi-task information fusion method for object detection pruning. However, such layer-by-layer pruning methods compute their importance measures locally, which do not consider the joint effect of multiple layers. This could affect their model performance when multiple layers are pruned simultaneously. In this paper, we introduce a globally computed saliency measure for visual detection pruning, which is defined as the gradients of the detection utility with respect to channel features.

\subsection{Visual Saliency Methods for CNNs}

Saliency methods are widely used to explain the predictions of deep neural networks by locating the regions that are relevant to a certain image's classification. Since the introduction of saliency maps~\cite{Simonyan14} in deep learning, several saliency methods for convolutional neural networks (CNNs) have been proposed~\cite{Springenberg15,Zhou2016,Selvaraju2017}. In their methods, saliency maps are heat maps that aim to provide insight into which image pixels contribute the most to the prediction of a network. However, those methods are not class-discriminative. Class Activation Map (CAM)~\cite{Zhou2016} is a class-discriminative classification saliency that allows seeing the class that a CNN predicts and the regions in the image that the network is interested in for predicting that class. However, it can only be applied to CNN architectures that perform global average pooling over the feature maps right before prediction. To address the drawbacks of CAM, Grad-CAM~\cite{Selvaraju2017}, a generalization of CAM, is introduced to be compatible with general neural architectures. However, Grad-CAM cannot localize multiple occurrences of an object in an image. Also, it cannot completely capture the entire object. Grad-CAM++~\cite{Chattopadhay2018} is an enhanced version of Grad-CAM. It addresses the limitations of Grad-CAM by using a weighted average of pixel-wise gradients. The CAM family is designed for classification tasks, and it aims to highlight class-wise attention. In this paper, we propose a novel gradient-based saliency measure for visual detection. To the best of our knowledge, this is the first gradient-based detection saliency that evaluates the contribution of each channel to the final detection. We leverage it to prune visual detectors for efficient autonomous driving perception.

\begin{figure*}[!t]
\centering
\begin{tabular}{cccc}
  \toprule
  Original Sample & Method & Ch. \#61 & Ch. \#356\\
  \midrule
  \multirow{-2}{*}{
  \includegraphics[width=.23\linewidth, height=.7in]{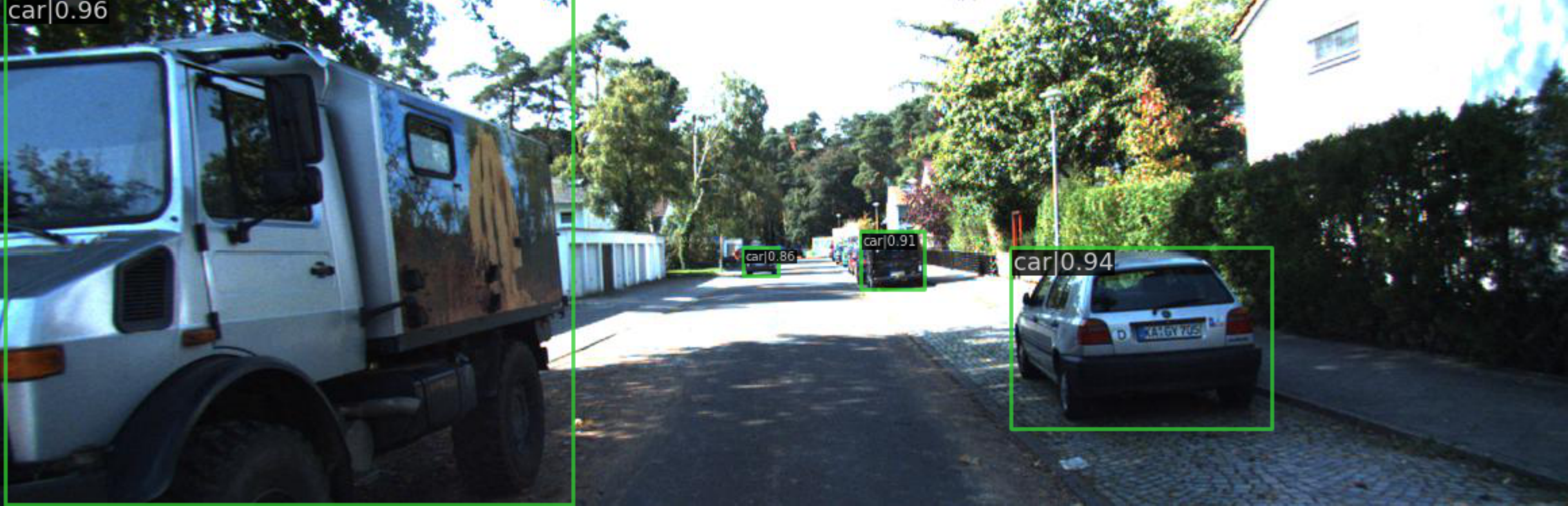}} &
  \multirow{2}{*}[3em]{Without Reweighting} &
  \includegraphics[width=.23\linewidth, height=.7in]{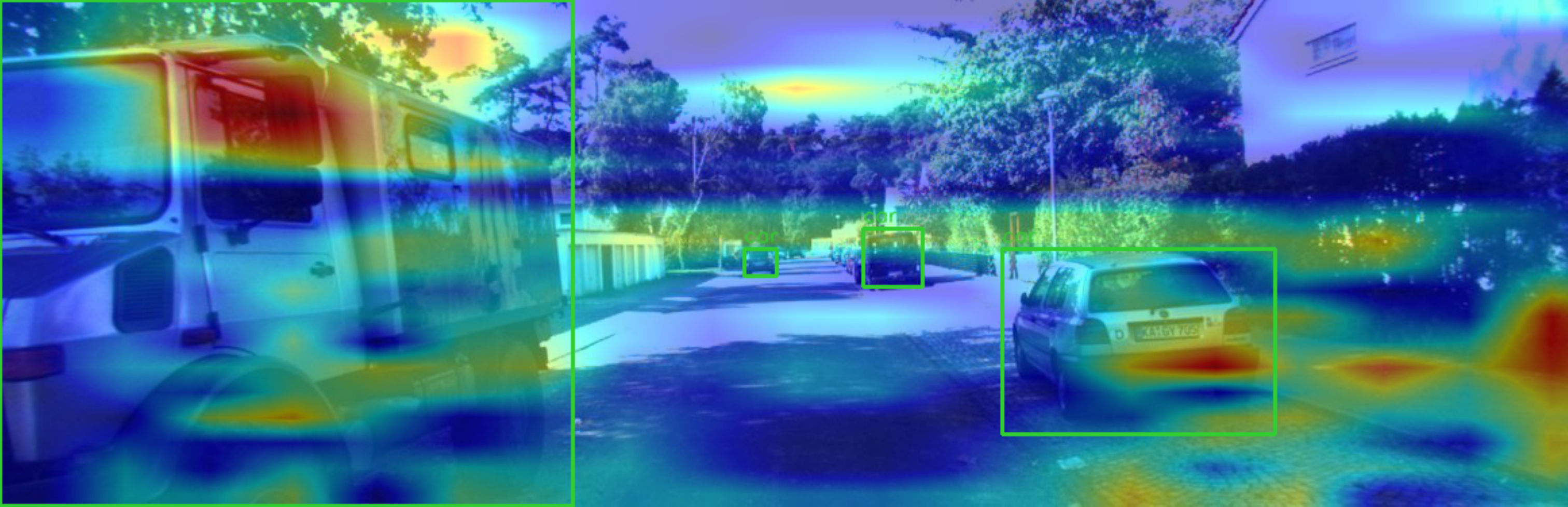} &
  \includegraphics[width=.23\linewidth, height=.7in]{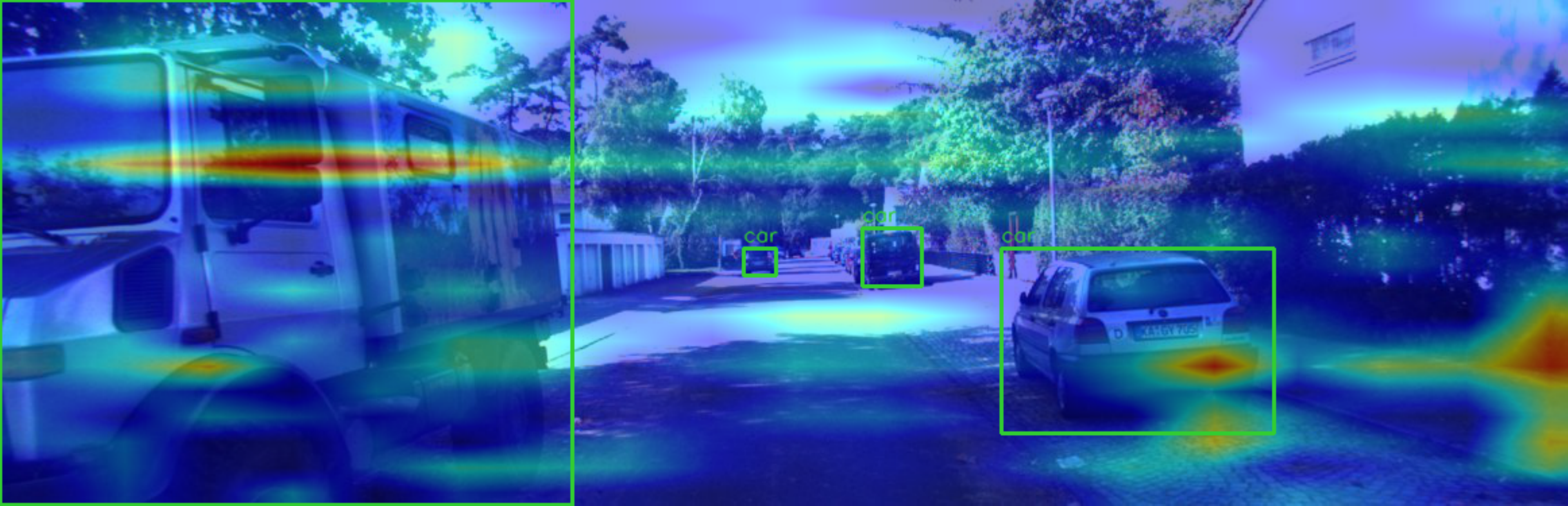}\\
  \cmidrule{2-4}
  & \multirow{2}{*}[3em]{With Reweighting} &
  \includegraphics[width=.23\linewidth, height=.7in]{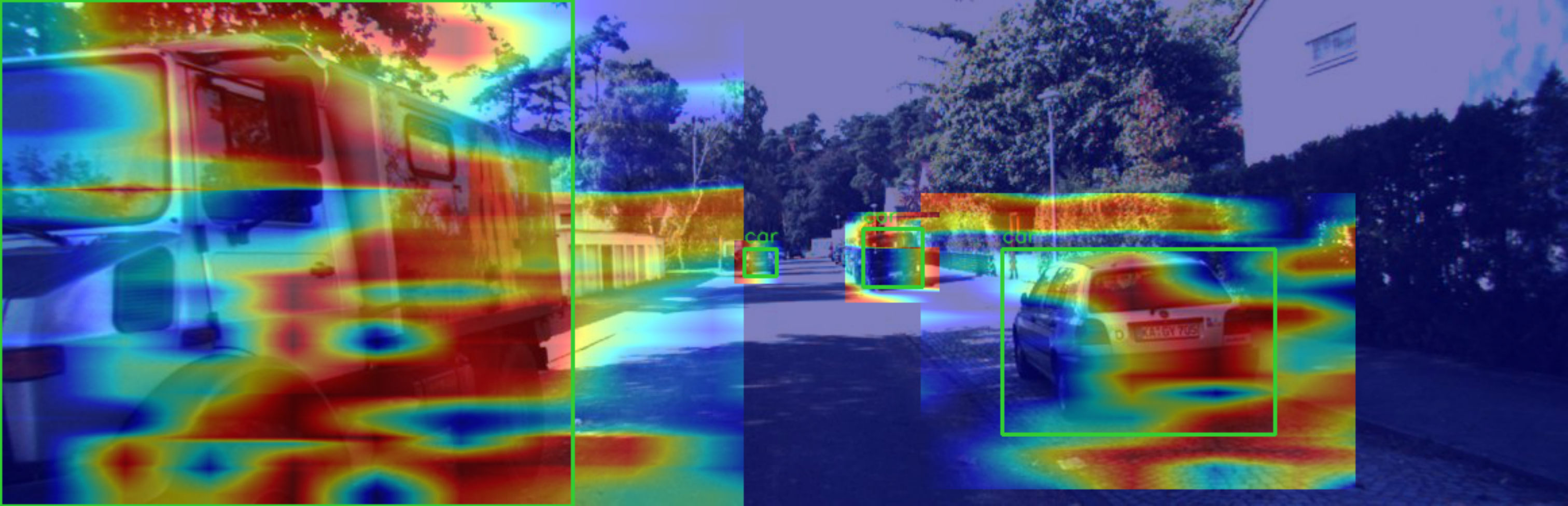} &
  \includegraphics[width=.23\linewidth, height=.7in]{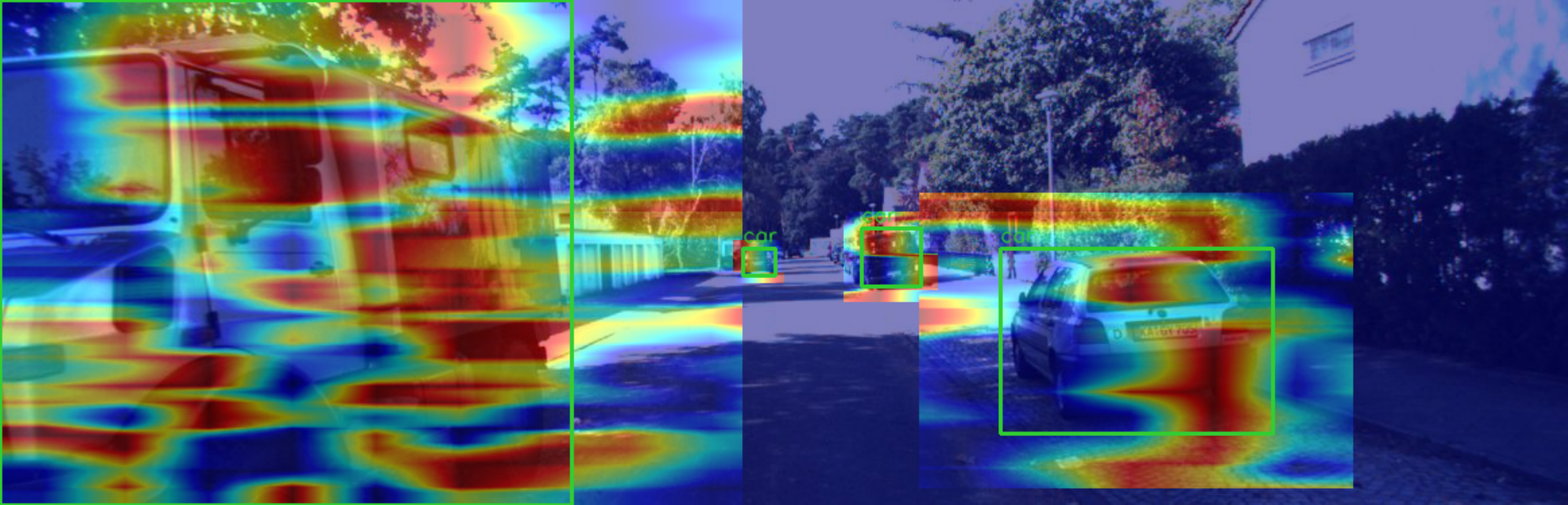}\\
  \bottomrule 
\end{tabular}
\caption{Saliency maps without and with our bounding-box-aware reweighting on a KITTI example. The saliency maps were generated from the YOLOX-S neck.bottom\_up\_blocks.1.final\_conv layer. Best viewed when zoomed in.}
\label{fig:reweighting results}
\end{figure*}

\section{Methodology}
\label{method}

Unlike filter pruning approaches which directly investigate the filters, channel pruning focuses on features produced by filters. It discards unimportant feature maps and their corresponding filters based on a certain importance measure.
Unlike many local pruning approaches, we argue that the importance measure of a channel should consider the channel's contribution to the final prediction.
In this paper, we propose a novel gradient-based saliency measure for visual detection tasks and use it to guide our pruning.
We define the visual saliency of the $k$th channel in layer $l$ as $w_k^l$:
\begin{equation} \label{eq:weight}
w_k^l = \sum_{i}\sum_{j}\beta_{(i,j)}^{kl}\cdot relu(\frac{\delta \mathcal{L}_{det}}{\delta A_{(i,j)}^{kl}})
\end{equation} 

\noindent where $\mathcal{L}_{det}$ is the overall loss for the object detection task, $A_{(i,j)}^{kl}$ is the feature point at location $(i,j)$ in the $k$th feature map of layer $l$, $\beta_{(i,j)}^{kl}$ is the corresponding reweighting coefficient, and $relu$ is the Rectified Linear Unit activation function. 
The overall detection loss $\mathcal{L}_{det}$ consists of the classification loss $\mathcal{L}_{cls}$ and bounding box regression loss $\mathcal{L}_{box}$: 
\begin{equation} \label{loss}
\mathcal{L}_{det} = \sum_{m=1}^M\mathbb{I}_m^{obj} (\lambda_{cls}\mathcal{L}_{cls}+\lambda_{box}\mathcal{L}_{box})
\end{equation}
where $\mathbb{I}_m^{obj}$ represents whether the $m$-th bounding box contains an object (i.e., $\mathbb{I}_m^{obj}$ = 1) or not (i.e., $\mathbb{I}_m^{obj}$ = 0) and $M$ is the number of predicted bounding boxes. $\lambda_{cls}$ and $\lambda_{box}$ are hyperparameters to balance the two loss terms' contribution. 

To filter out irrelevant features, we incorporate ground truth bounding box information. We define the reweighting coefficient $\beta_{(i,j)}^{kl}$ as:
\begin{equation} \label{eq:beta}
\beta_{(i,j)}^{kl}=
\begin{cases}
1\qquad\qquad\quad if\; (i, j)\; \in area(\mathcal{B})\\
f_{decay}(i,j) \quad else\; if\; (i, j)\; \in neighbor(\mathcal{B})\\
0 \qquad\qquad\quad otherwise
\end{cases}
\end{equation}

\noindent where $\mathcal{B}$ is a set of the ground truth bounding boxes and $neighbor(\mathcal{B})$ is a set of the surrounding pixels around the ground truth bounding boxes with the margins in proportion to the bounding boxes' sizes. The margin ratio is empirically set. $f_{decay}(\cdot,\cdot)$ is a decay function, which is defined as: 
\begin{equation} \label{eq:decay}
f_{decay}(i, j)=a\{(i-i_c)^2+(j-j_c)^2\}^{-s}
\end{equation}
where $a$ is a non-zero constant, $(i_c,j_c)$ is the center of the corresponding ground truth bounding box, and $s$ is a real number. Instead of focusing on only pixels that are inside the ground truth boxes, we relax every ground truth box in proportion
to its size to obtain more information on the regions surrounding the box. The reason why we need to consider the neighboring regions is twofold: ($i$) surrounding context information can be useful in detecting the object; ($ii$) the neighborhood information diminishes gradually from the bounding box center (not a hard cut-off at the bounding box boundaries).
Fig.~\ref{fig:reweighting results} shows the visual effects of adding the reweighting coefficient.

For a given input sample, the importance of the $k$th channel of layer $l$ is defined as:
\begin{equation} \label{eq:total_score}
S^{kl} = \sum_{i}\sum_{j}norm(relu(w_k^l \cdot A_{(i,j)}^{kl}))
\end{equation}
where $norm$ is a min-max normalization function to rescale the saliency inside the relaxed bounding boxes.
Algorithm~\ref{alg:algorithm} summarizes the overall scheme for the proposed saliency-guided channel pruning algorithm for a particular layer $l$. 

\begin{algorithm}[htb]
    \caption{Saliency-Guided Channel Pruning for the $l$-th Layer}
    \label{alg:algorithm}
    \textbf{Input:} Pre-trained\_model, Pruning\_rate $r$, $N$ input\_samples, $K$ channels in the layer $l$\\
    \textbf{Output:} Pruned\_model
    \begin{algorithmic}[1] 
        \FOR{each input sample}
            \FOR{k=1 \textbf{to} $K$} 
                \STATE Compute the channel saliency:\\
                $w_k^l = \sum_{i}\sum_{j}\beta_{(i,j)}^{kl}\cdot relu(\frac{\delta \mathcal{L}_{det}}{\delta A_{(i,j)}^{kl}})$
                \STATE Compute the channel importance:\\
                $S^{kl} = \sum_{i}\sum_{j}norm(relu(w_k^l \cdot A_{(i,j)}^{kl}))$
            \ENDFOR
        \ENDFOR
        \STATE Average:
        $\{S_{avg}^{kl}\}_{k=1}^K = \{\frac{\sum_{n=1}^NS_n^{kl}}{N}\}_{k=1}^K$
        \STATE Sort the mean channel-wise importance in ascending order
        \STATE Prune the lowest channels and their corresponding filters by pruning\_rate $r$
        \STATE Fine-tune the pruned\_model
    \end{algorithmic}
\end{algorithm}

\section{Experiments}
\label{experiments}

\subsection{Experimental Setup}

We use the YOLOX-S~\cite{Ge2021} and YOLOF~\cite{Chen2021}, state-of-the-art variants in the YOLO family, as the base models in our experiments. CSPDarkNet53~\cite{Redmon2013} and ResNet50~\cite{He2016} are used as the backbones for YOLOX-S and YOLOF, respectively. Both base models are pre-trained on MS-COCO 2017~\cite{Lin2014} dataset. In our experiments, we use the KITTI~\cite{Geiger2012} and COCO\_traffic datasets~\cite{Qizhen2022}. For the KITTI dataset, we follow the same convention for combining the categories and splitting the dataset as in~\cite{Choi2022}. To be specific, there are three categories for the KITTI dataset: car, cyclist, and pedestrian. We use the first half of the 7,481 training images (sorted by the image IDs in ascending order) as the training set and the second half as the validation set (the testing images do not have labels). For the COCO\_traffic dataset, we select 13 categories related to autonomous driving from MS-COCO 2017 dataset~\cite{Lin2014} (i.e., person, bicycle, car, motorcycle, bus, train, truck, traffic light, fire hydrant, stop sign, parking meter, dog, and cat). A total of 25,180 training images and 1,092 testing images are used.

\begin{figure}[t]
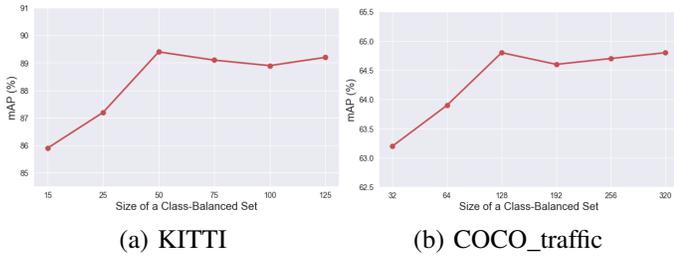

\begin{minipage}[b]{.49\linewidth}
  \centering
  \centerline{\includegraphics[width=4.5cm]{figures/size1.pdf}}
  \centerline{(a) KITTI}\medskip
\end{minipage}
\hfill
\begin{minipage}[b]{0.49\linewidth}
  \centering
  \centerline{\includegraphics[width=4.5cm]{figures/size2.pdf}}
  \centerline{(b) COCO\_traffic }\medskip
\end{minipage}
\caption{mAP comparison of YOLOX-S models using different sizes of a class-balanced sample set with a pruning rate of 0.3 on (a) KITTI and (b) COCO\_traffic datasets. }
\label{fig:bt_sizes}
\end{figure}

We observe that both datasets are class-imbalanced (e.g., the number of cars is much larger than the numbers for other classes). Since our approach computes the task gradients with respect to features, it is a data-driven approach. Ideally, such a data-driven method requires a large number of input images for its reliability, but it is typically not feasible in practice due to intensive computational cost. Thus, we use a subset of input images, and in order to avoid bias in computing the average channel saliency due to imbalanced datasets, we carefully select each set of input images. For each class-balanced set of samples, we compute the channel detection saliency, perform channel pruning based on the saliency measure and evaluate the performance of the pruned model on the testing set. Fig.~\ref{fig:bt_sizes} shows the empirical results. Each performance graph almost reaches a plateau and has equally good results after its peak. Based on this observation, we use a class-balanced set of 50 training images for the KITTI and 128 training images for the COCO\_traffic dataset to calculate the average channel importance. After performing the saliency-guided channel pruning with various pruning rates (where the pruning rate refers to the percentage of channels discarded in a layer), we fine-tune the pruned models with the SGD optimizer. Specifically, we perform the fine-tuning for 300 epochs on the KITTI dataset and 100 epochs on the COCO\_traffic dataset with a batch size of 8, momentum of 0.9, weight decay of 0.0005, and initial learning rate of 0.01. For both datasets, we use the mean Average Precision (mAP) with IoU threshold of 0.5 as our overall performance metric. In addition, APs calculated across different object scales (i.e., small, medium, and large objects) with IoU threshold of 0.5 are used for both datasets. AP-s, AP-m, and AP-l are defined as the AP for small (area less than $32^2$, medium (area between $32^2$ and $96^2$), and large (area than $96^2$) objects, respectively.

\subsection{Pruning Rates vs. mAP}

In this subsection, we evaluate the YOLOX-S and YOLOF models pruned by our method with various pruning rates on both the KITTI and COCO\_traffic datasets. 

\subsubsection{Results on KITTI}

\begin{table}[htb]
    \centering
    \addtolength{\tabcolsep}{-4pt}
    \begin{tabular}{lccccccc}
        \toprule
        \multirow{2}{*}{Method} & Pruning & Flops & Params & AP-s & AP-m & AP-l & mAP \\
         & Rate & (G) & (M) & (\%) & (\%) & (\%) & (\%)\\
        \midrule
        Vanilla YOLOX-S & 0.0 & 26.65 & 8.94 & 75.4 & 89.4 & 92.8 & 87.3 \\
        \midrule
        L1 Pruning~\cite{Li2017} 
        & \multirow{4}{*}{0.3} & 18.71 & 5.40 & 78.7 & 90.0 & \textbf{93.4} & 88.5\\
        Net Slimming~\cite{Liu2017} 
        &  & 16.49 & 5.96 & 68.3 & 86.3 & 93.1 & 84.0 \\
        CHIP~\cite{Sui2021} &  & 18.88 & 5.35 & 79.9 & \textbf{90.3} & 92.7 & 88.9 \\
        \textbf{Ours} & & 18.88 & 5.35 & \textbf{81.1} & 90.1 & 93.1 & \textbf{89.1}  \\
        \midrule
        L1 Pruning~\cite{Li2017} & \multirow{4}{*}{0.5} & 14.99 & 3.88 & 65.9 & 83.1 & 89.6 & 81.1\\
        Net Slimming~\cite{Liu2017} &  & 14.22 & 4.08 & 57.8 & 80.0 & 91.2 & 77.8 \\
        CHIP~\cite{Sui2021} &  & 15.01 & 3.61 & 74.1 & \textbf{88.1} & 92.3 & 86.2 \\
        \textbf{Ours} & & 15.01 & 3.61 & \textbf{75.1} & \textbf{88.1} & \textbf{92.6} &\textbf{86.6} \\
        \midrule
        L1 Pruning~\cite{Li2017} & \multirow{4}{*}{0.7} & 12.36 & 2.22 & 24.1 & 33.6 & 48.6 & 34.9 \\
        Net Slimming~\cite{Liu2017} & & 12.69 & 2.66 & 44.7 & 67.9 & 84.7 & 66.7 \\        
        CHIP~\cite{Sui2021} &  & 12.50 & 2.47 & 64.0 & \textbf{83.2} & 91.4 & 80.9 \\
        \textbf{Ours} & & 12.50 & 2.47 & \textbf{64.4} & 83.1 & \textbf{91.6} & \textbf{81.1} \\
        \bottomrule
    \end{tabular}
    \caption{Comparison of pruned YOLOX-S models on KITTI.}
    \label{tab:YOLOX-S_KITTI}
\end{table}

\begin{table}[htb]
    \centering
    \addtolength{\tabcolsep}{-4pt}
    \begin{tabular}{lccccccc}
        \toprule
        \multirow{2}{*}{Method} & Pruning & Flops & Params & AP-s & AP-m & AP-l & mAP \\
         & Rate & (G) & (M) & (\%) & (\%) & (\%) & (\%)\\
        \midrule
        Vanilla YOLOF & 0.0 & 106.90 & 42.39 & 74.4 & 89.4 & 93.6 & 87.4 \\
        \midrule
        L1 Pruning~\cite{Li2017} 
        & \multirow{4}{*}{0.3} & 63.09 & 27.87 & 65.1 & 80.1 & 84.0 & 77.8 \\
        Net Slimming~\cite{Liu2017} 
        & & 80.54 & 25.54 & 67.3 & 84.5 & 89.4 & 81.8\\     
        CHIP~\cite{Sui2021} &  & 60.83 & 26.75 & 74.1 & 88.2 & 92.4 & 86.3 \\
        \textbf{Ours} & & 60.83 & 26.75 & \textbf{76.6} & \textbf{88.8} & \textbf{92.7} & \textbf{87.2} \\
        \midrule
        L1 Pruning~\cite{Li2017} 
        & \multirow{4}{*}{0.5} & 50.83 & 22.34 & $<$10 & $<$10 & $<$10 & $<$10 \\
        Net Slimming~\cite{Liu2017} 
        &  & 45.89 & 18.04 & 62.1 & 81.7 & 85.9 & 78.3 \\    
        CHIP~\cite{Sui2021} &  & 38.34 & 19.02 & 67.7 & 85.2 & 90.3 & 82.5 \\
        \textbf{Ours} & & 38.34 & 19.02 & \textbf{70.6} & \textbf{88.2} & \textbf{91.3} & \textbf{85.0} \\
        \midrule
        L1 Pruning~\cite{Li2017} 
        & \multirow{4}{*}{0.7} & 39.77 & 17.32 & $<$10 & $<$10 & $<$10 & $<$10 \\
        Net Slimming~\cite{Liu2017} 
        &  & 27.29 & 15.22 & 41.2 & 59.9 & 65.3 & 57.4\\    
        CHIP~\cite{Sui2021} &  & 23.23 & 13.61 & 61.5 & 80.4 & 86.3 & 77.5 \\
        \textbf{Ours} & & 23.23 & 13.61 & \textbf{65.2} & \textbf{82.5} & \textbf{89.0} & \textbf{80.0} \\
        \bottomrule
    \end{tabular}
    \caption{Comparison of pruned YOLOF models on KITTI.}
    \label{tab:YOLOF_KITTI}
\end{table}

We apply our pruning approach to both baseline detectors on the KITTI dataset and compare its pruning performance with other state-of-the-art pruning methods, i.e., $L$1-norm based pruning~\cite{Li2017}, Network Slimming~\cite{Liu2017}, and CHIP~\cite{Sui2021}. The experimental results for YOLOX-S and YOLOF are shown in TABLE~\ref{tab:YOLOX-S_KITTI} and TABLE~\ref{tab:YOLOF_KITTI}, respectively. It is observed that on the KITTI dataset, the proposed method achieves the best performance among the methods at all pruning rates. Moreover, in TABLE~\ref{tab:YOLOX-S_KITTI}, we can see that our approach can even achieve an improvement of 1.8\% mAP over the vanilla/unpruned model with 40.2\% parameters (29.2\% FLOPs) pruned. In TABLE~\ref{tab:YOLOF_KITTI}, our method can maintain a comparable performance (0.2\% mAP drop) even after reducing 36.9\% parameters (43.1\% FLOPs). Furthermore, our approach outperforms the state-of-the-art competing methods in detecting small objects at all pruning rates for both base models.

\subsubsection{Results on COCO\_traffic}

\begin{figure*}[!t]
\centering
\begin{tabular}{@{} ccccc @{}}
  \toprule
  \multirow{2}{*}[3em]{CHIP~\cite{Sui2021}} & \includegraphics[width=.2\linewidth, height=.8in]{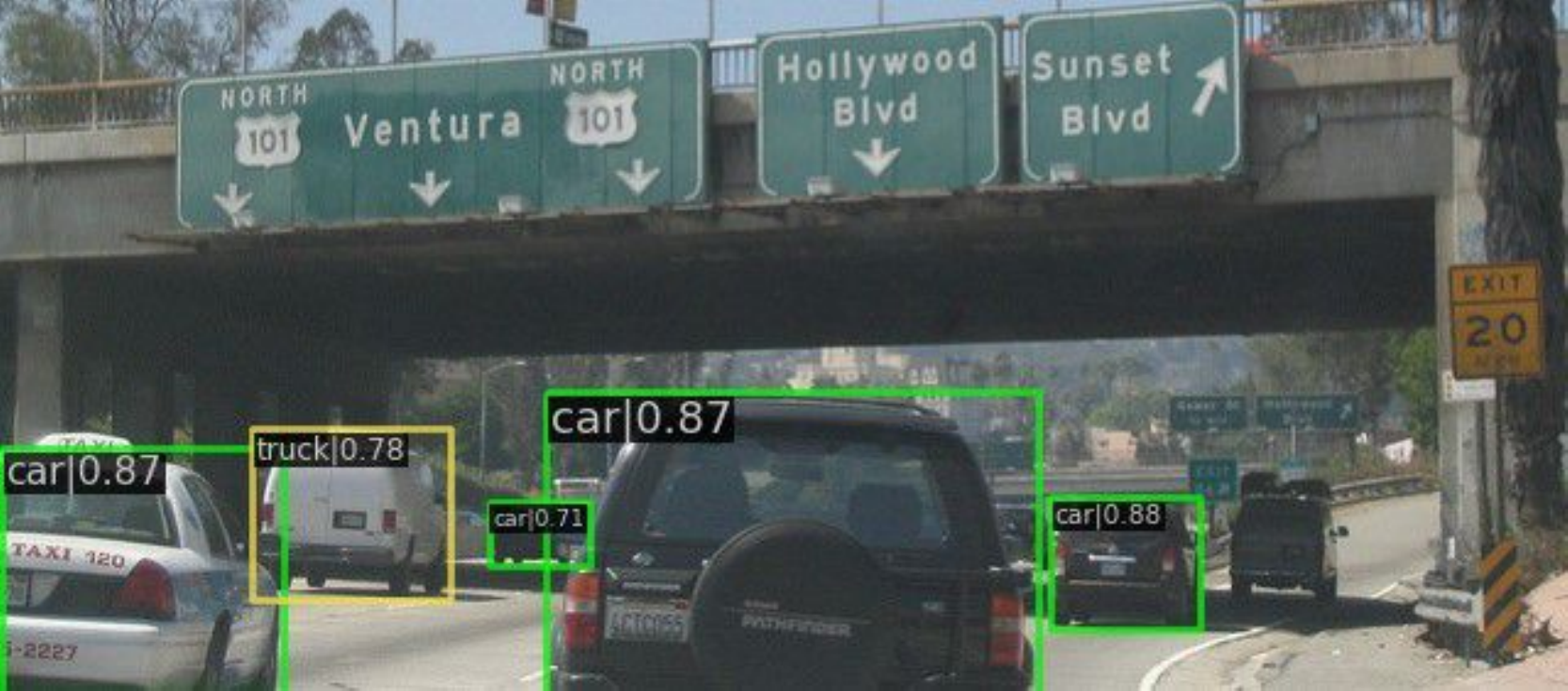} &
  \includegraphics[width=.2\linewidth, height=.8in]{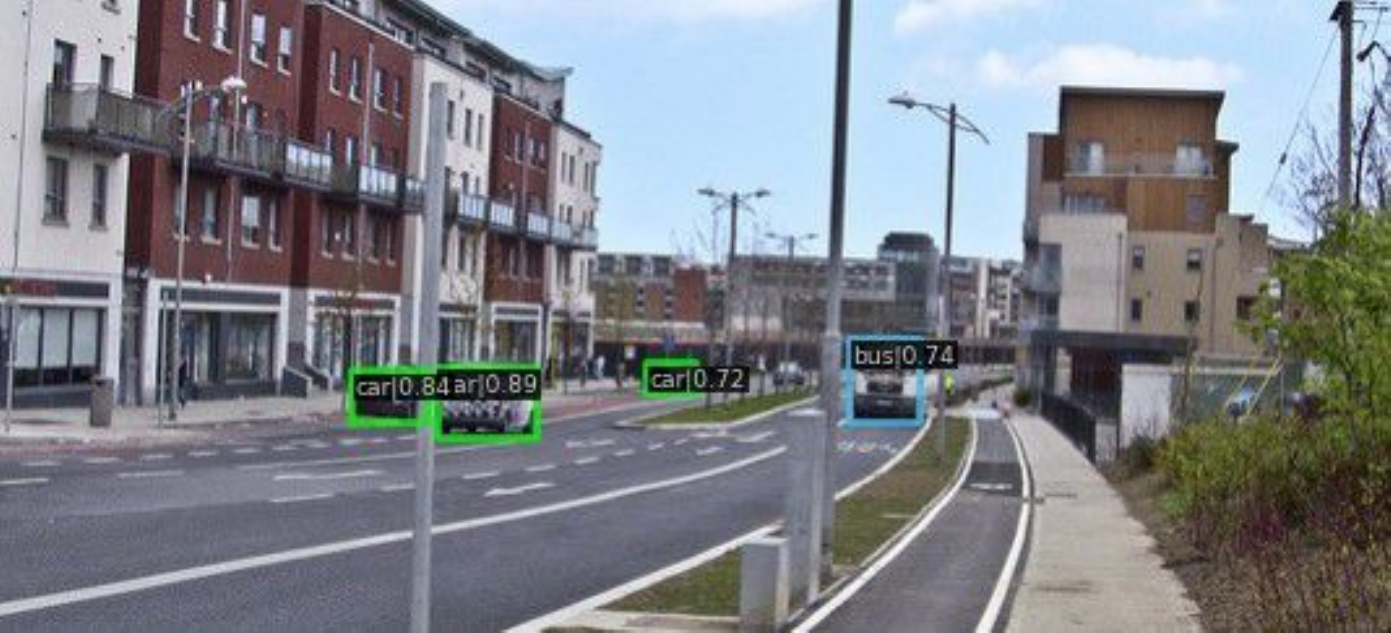} &
  \includegraphics[width=.2\linewidth, height=.8in]{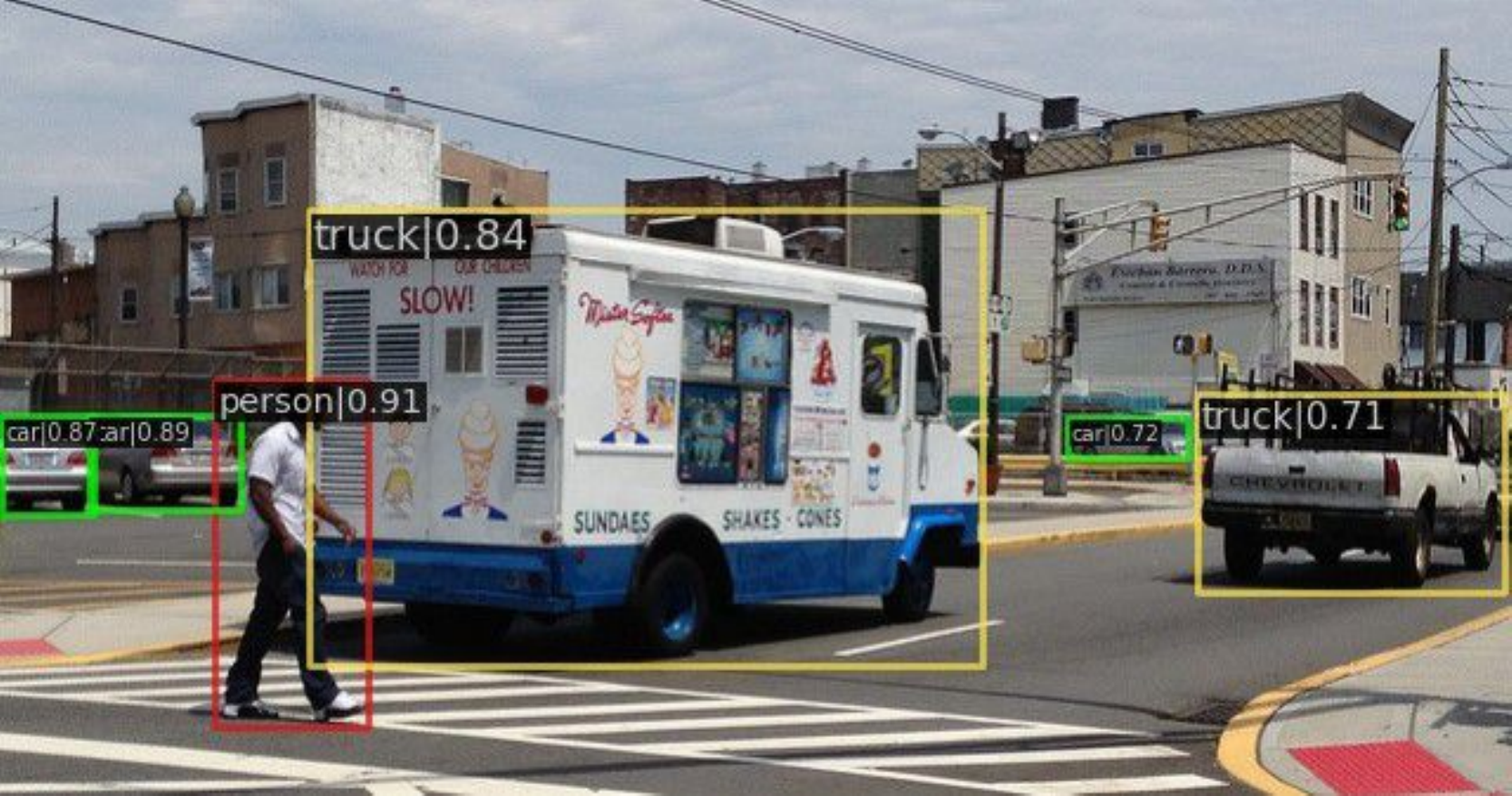} &
  \includegraphics[width=.2\linewidth, height=.8in]{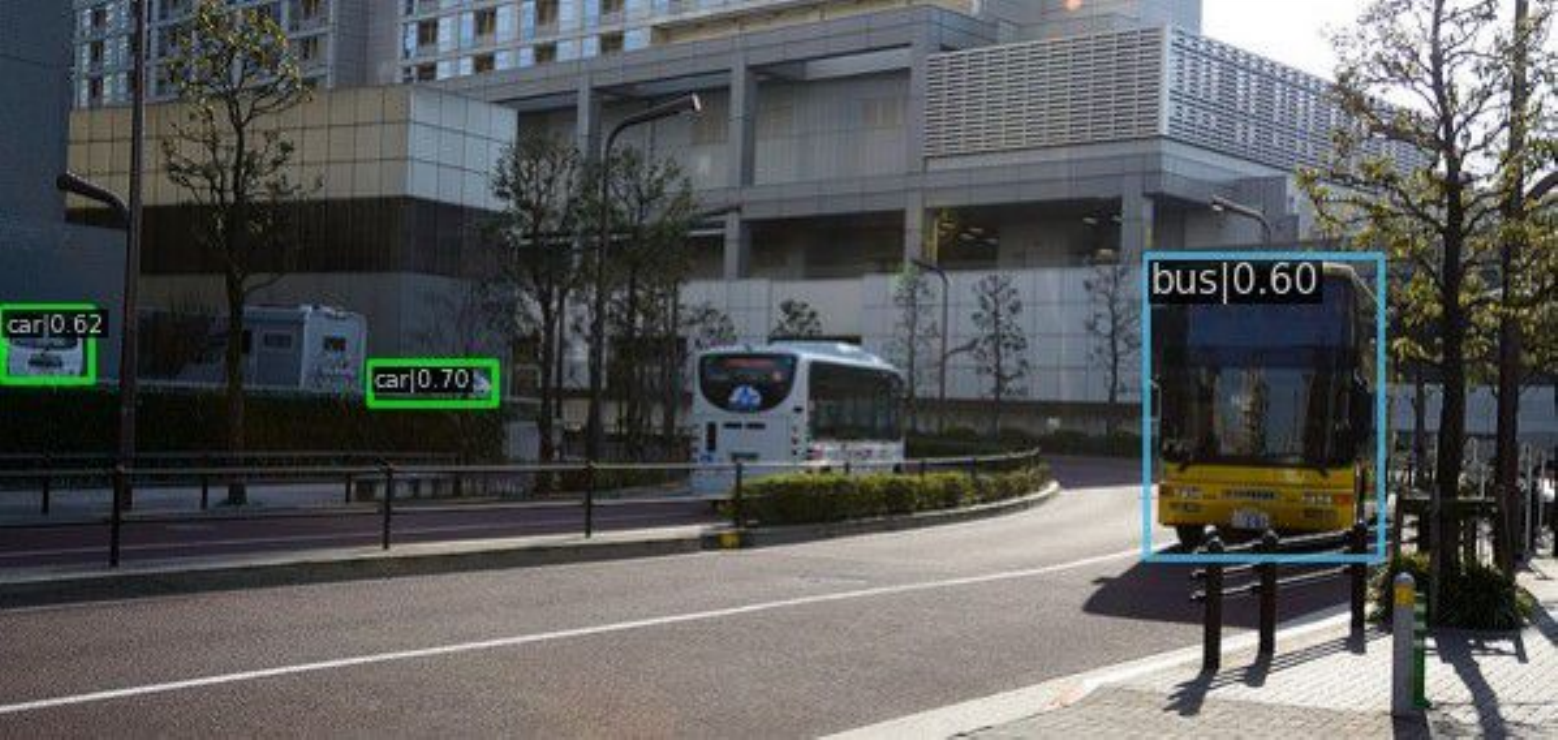}\\
  \midrule
  \multirow{2}{*}[3em]{Ours} & \includegraphics[width=.2\linewidth, height=.8in]{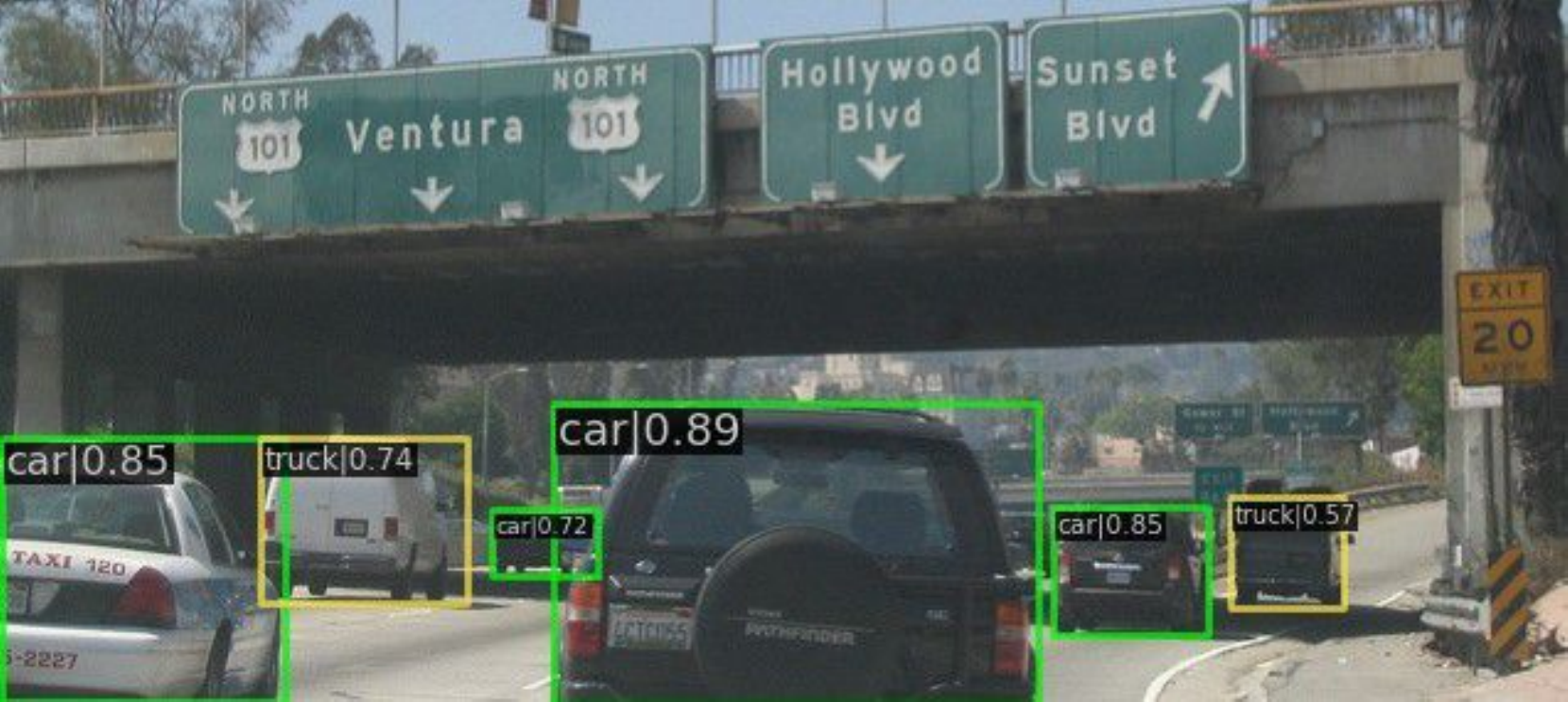} &
  \includegraphics[width=.2\linewidth, height=.8in]{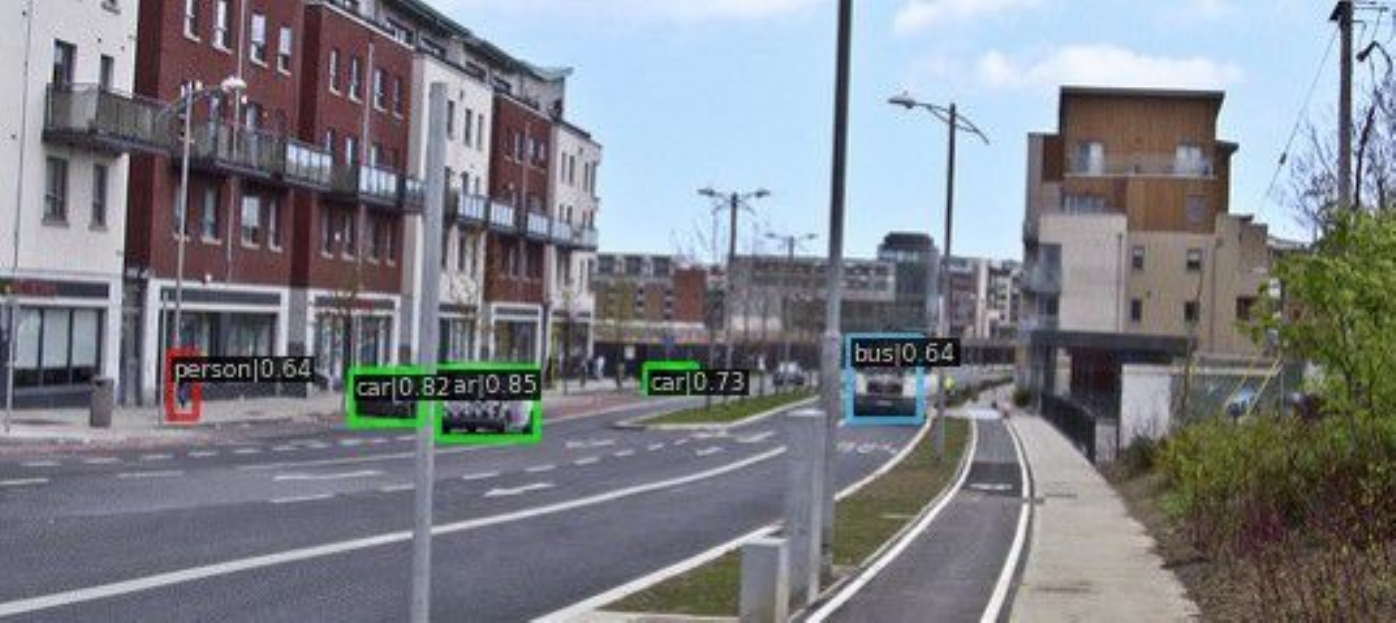} &
  \includegraphics[width=.2\linewidth, height=.8in]{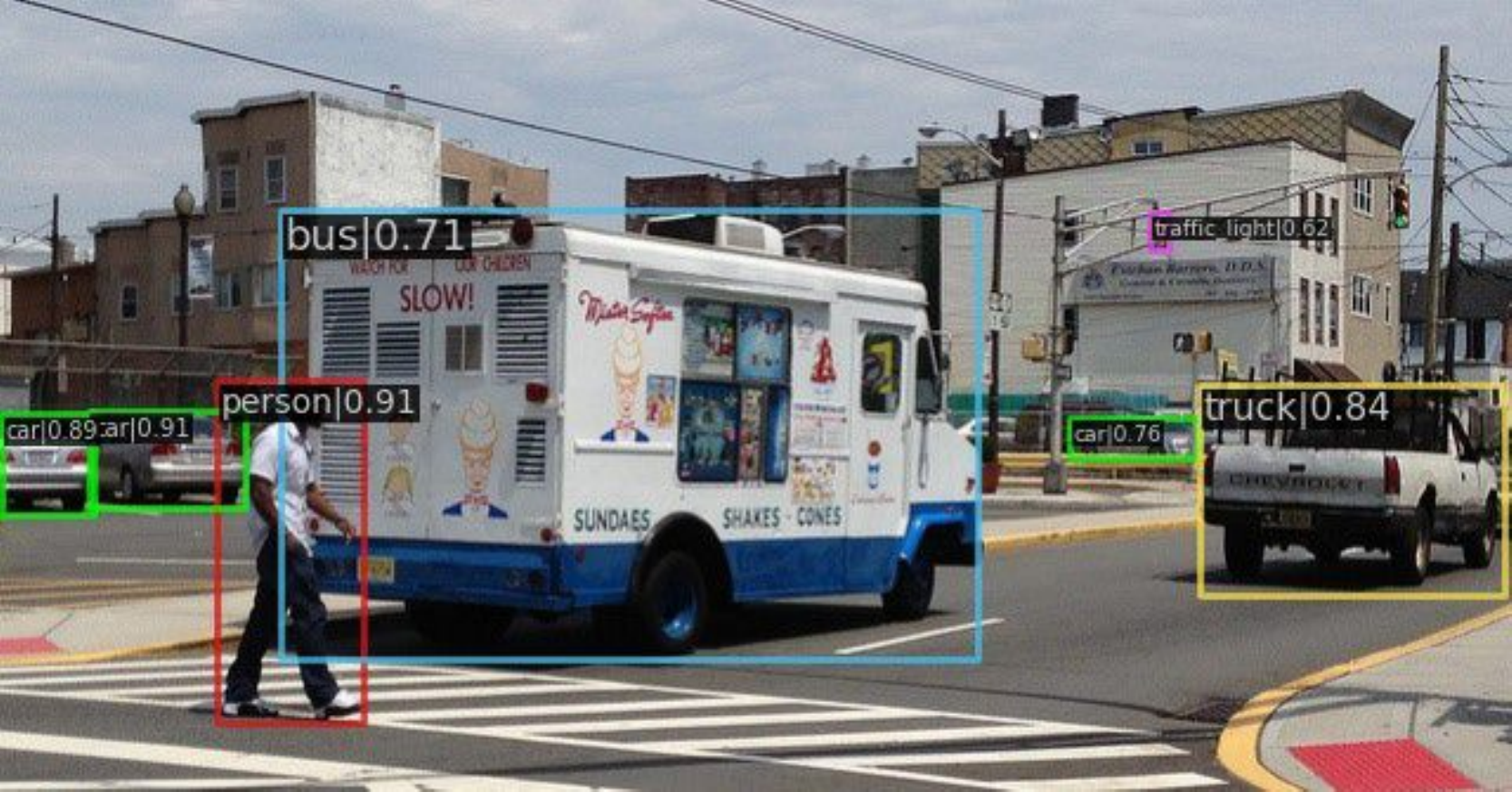} &
  \includegraphics[width=.2\linewidth, height=.8in]{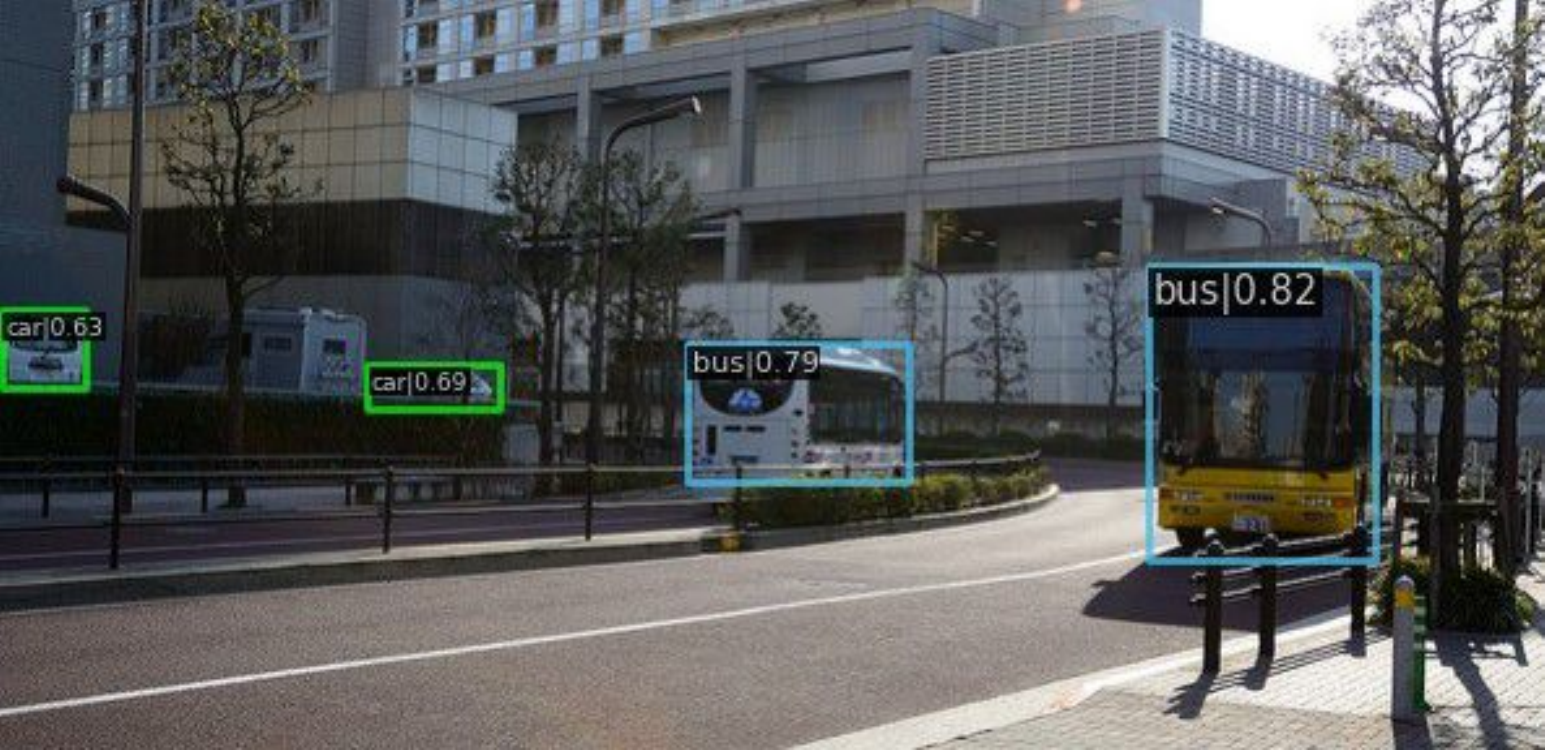}\\ 
  \bottomrule
\end{tabular}
\caption{Visual comparison between CHIP~\cite{Sui2021} and ours on COCO\_traffic examples. The detection results were from YOLOX-S models with a pruning rate of 0.3. Best viewed when zoomed in.}
\label{fig:qualitative}
\end{figure*}

On the COCO\_traffic dataset, the results of the pruned YOLOX-S and YOLOF models are shown in TABLE~\ref{tab:YOLOX-S_COCO} and TABLE~\ref{tab:YOLOF_COCO}, respectively. We can find that the proposed approach gives superior performance to other state-of-the-art methods on all the pruned YOLOX-S models. For pruned YOLOF models, our approach gives comparable performance at the pruning rate of 0.1 and better performance at larger pruning rates than the other competing approaches. In addition, our approach outperforms the state-of-the-art methods in detecting small objects at all pruning rates. For example, our pruned YOLOX-S model can achieve 9.8\% higher mAP (on average) for small-scale objects than the models using other pruning methods with 50.7\% and 36.7\% model size and FLOPs reductions, respectively. Visual comparison between CHIP~\cite{Sui2021} and our approach on COCO\_traffic examples is shown in Fig.~\ref{fig:qualitative}. Our approach demonstrates better detection results than the state-of-the-art method, especially in small-scale boxes. One possible reason is that the relaxed areas of small-scale bounding boxes carry more useful information about the surrounding context than of other-scale boxes, and our reweighting strategy incorporating the surrounding context information of ground truth boxes can capture that important information. 

\begin{table}[htp]
    \centering
    \addtolength{\tabcolsep}{-4pt}
    \begin{tabular}{lccccccc}
        \toprule
        \multirow{2}{*}{Method} & Pruning & Flops & Params & AP-s & AP-m & AP-l & mAP \\
        & Rate & (G) & (M) & (\%) & (\%) & (\%) & (\%)  \\
        \midrule
        Vanilla YOLOX-S & 0.0 & 26.67 & 8.94 & 44.6 & 72.5 & 83.9 & 68.7\\
        \midrule
        L1 Pruning~\cite{Li2017}  
        & \multirow{4}{*}{0.2} & 21.23 & 6.61 & 42.5 & 71.8 & \textbf{81.1} & \textbf{66.0} \\
        Net Slimming~\cite{Liu2017} & & 19.48 & 7.08 & 38.1 & 67.4 & 79.6 &  62.2 \\ 
        CHIP~\cite{Sui2021} & & 21.29 & 6.41 & 41.2 & \textbf{71.9} & 79.9 & 65.8 \\
        \textbf{Ours} & & 21.29 & 6.41 & \textbf{46.0} & 69.2 & 80.6 & 65.9 \\
        \midrule
        L1 Pruning~\cite{Li2017} & \multirow{4}{*}{0.4} &  16.48 & 4.78 & 25.9 & 56.1 & 68.2 &  51.1  \\
        Net Slimming~\cite{Liu2017} & & 16.00 & 4.96 & 30.6 & 63.5 & 5.9 & 57.6 \\
        CHIP~\cite{Sui2021} &  & 16.87 & 4.41 & 37.5 & 66.7 & \textbf{78.4} & 62.5 \\
        \textbf{Ours} & & 16.87 & 4.41 & \textbf{41.1} & \textbf{67.5} & 78.1 & \textbf{62.9} \\
        \midrule
        L1 Pruning~\cite{Li2017} & \multirow{4}{*}{0.6} &  12.64 & 2.56 & $<$10 & 15.9 & 30.1 &  18.4 \\
        Net Slimming~\cite{Liu2017} & & 13.90 & 3.37 &25.2 & 54.5 & 73.2 & 51.0 \\
        CHIP~\cite{Sui2021} &  & 13.64 & 2.98 & 31.5 & \textbf{62.6} & 74.7 & 56.8 \\
        \textbf{Ours} & & 13.64 & 2.98 & \textbf{32.0} & 61.0 & \textbf{75.6} & \textbf{57.0} \\
        \bottomrule
    \end{tabular}
    \caption{Results of YOLOX-S models on COCO\_traffic.}
    \label{tab:YOLOX-S_COCO}
\end{table}

\begin{table}[htb]
    \centering
    \addtolength{\tabcolsep}{-4pt}
    \begin{tabular}{lccccccc}
        \toprule
        \multirow{2}{*}{Method} & Pruning & Flops & Params & AP-s & AP-m & AP-l & mAP \\
        & Rate & (G) & (M) & (\%) & (\%) & (\%) & (\%)\\
        \midrule
        Vanilla YOLOF & 0.0 & 107.15 & 42.39 & 42.9 & 71.0 & 84.5 & 67.0 \\
        \midrule
        L1 Pruning~\cite{Li2017} & \multirow{4}{*}{0.1} & 86.56 & 41.11 & 23.1 & 60.4 & 79.4 & 56.0 \\
        Net Slimming~\cite{Liu2017}& & 94.83 & 35.40 & 38.7 & 68.1 & 82.7 & 64.7 \\
        CHIP~\cite{Sui2021} & & 90.40 & 36.65 & 38.9 & 69.7 & \textbf{83.7} & \textbf{65.8} \\
        \textbf{Ours} & & 90.40 & 36.65 & \textbf{42.7} & \textbf{70.0} & 82.4 & 65.7 \\
        \midrule
        L1 Pruning~\cite{Li2017} & \multirow{4}{*}{0.3} & 66.39 & 34.62 & $<$10 & $<$10 & $<$10 & $<$10\\
        Net Slimming~\cite{Liu2017} & & 57.17 & 25.49 & 34.5 & 65.3 & 79.7 & 61.2 \\
        CHIP~\cite{Sui2021} & & 61.09 & 26.75 & 35.5 & 66.2 & \textbf{81.5} & 61.8 \\
        \textbf{Ours} & & 61.09 & 26.75 & \textbf{38.0} & \textbf{67.0} & 80.2 & \textbf{62.3} \\
        \midrule
        L1 Pruning~\cite{Li2017} & \multirow{4}{*}{0.5} & 43.23 & 20.58 & $<$10 & $<$10 & $<$10 & $<$10 \\
        Net Slimming~\cite{Liu2017} & & 44.77 & 18.83 & $<$10 & 27.1 & 25.2 & 20.2 \\
        CHIP~\cite{Sui2021} & & 38.59 & 19.02 & 26.4 & 54.7 & 69.1 & 51.2 \\
        \textbf{Ours} & & 38.59 & 19.02 & \textbf{26.8} & \textbf{55.5} & \textbf{71.6} & \textbf{52.4} \\
        \bottomrule
    \end{tabular}
    \caption{Results of YOLOF models on COCO\_traffic.}
    \label{tab:YOLOF_COCO}
\end{table}

\subsection{Ablation Study}

\subsubsection{Contribution of Different Components} 

In this ablation study, we consider three components (i.e., the gradients of the detection utility, ground truth bounding boxes, and their surrounding context information) in our detection saliency measure. To see the effect of the three components on pruning performance, we ablate each component at a time, perform the channel pruning, and evaluate the performance of the pruned model. The experiment is conducted on YOLOX-S with a pruning rate of 0.3 on KITTI dataset and the results are shown in TABLE~\ref{tab:ablation}. It is observed that each component has a positive effect on the final performance and the model pruned by considering the three components achieves the best performance, which demonstrates the efficacy of our proposed approach.        

\begin{table}[htb]
    \centering
    \small
    \begin{tabular}{c|c|cccc}
        \toprule
        Detector & \multicolumn{5}{c}{YOLOX-S}\\
        \midrule
        \multirow{3}{*}{Modules} & Gradients & $\times$ &$\checkmark$ &$\checkmark$ & $\checkmark$ \\
        & GT bounding box & $\times$ & $\times$ & $\checkmark$ & $\checkmark$ \\
        & Context info & $\times$ & $\times$  & $\times$ & $\checkmark$ \\
        \midrule
        Results & mAP & 87.3 & 88.5 &  88.7 & \textbf{89.4}  \\
        \bottomrule
    \end{tabular}
    \caption{Impact of the reweighting strategy incorporating ground truth (GT) bounding boxes and their surrounding context information in the detection saliency measure. The experiment is performed on YOLOX-S models with a pruning rate of 0.3 on the KITTI dataset.}
    \label{tab:ablation}
\end{table}

\subsubsection{Decay Weighting Functions}
\label{decay}

\begin{figure}[htb]
    \centering
    \includegraphics[width=\linewidth, height=1.8in]{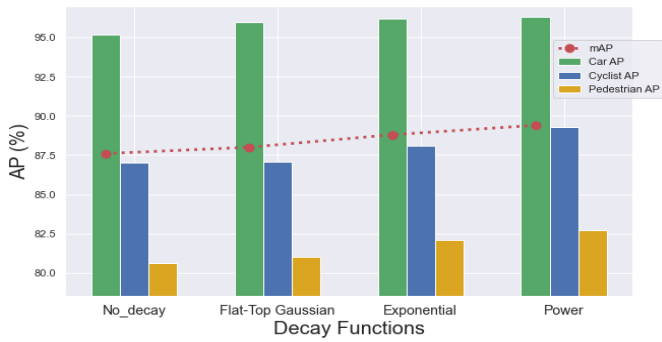}
    \caption{mAP comparison of decay functions applied in YOLOX-S with a pruning rate of 0.3 on the KITTI dataset}
    \label{fig:decay_func}
\end{figure}

To utilize more information on the neighboring pixels of the ground truth bounding boxes, we add margins to the boxes and apply a decay function to those pixels in the margins. As mentioned above, utilizing the surrounding context information in evaluating the detection saliency further improves the performance of the pruned model. Here we evaluate the model performance of alternative decay functions. Three decay functions (i.e., flat-top Gaussian function, exponential function, and power function) were considered for the reweighting approach. Each function is evaluated on YOLOX-S with a pruning rate of 0.3 on the KITTI dataset, and the results are shown in Fig.~\ref{fig:decay_func}. The model using a power decay function achieves the highest mAP while the model without a decay function gives the lowest one. Thus, we choose a power function as our decay function to obtain the neighboring context information of the ground truth bounding boxes.   

\section{Conclusion}
\label{conclusion}

In this paper, we have introduced a novel gradient-based channel saliency measure for visual detection tasks and have leveraged it to guide the pruning of deep visual detectors. Our pruning method achieves great pruning rates with little performance degradation on the KITTI and COCO\_traffic datasets using YOLOX and YOLOF as bases. On the KITTI dataset, our pruned model can even beat the original unpruned model by 1.8\% mAP with only 59.8\% parameters. In addition, we find that our method performs especially well on small-scale objects.

\section*{Acknowledgment}

This research was supported by the National Science Foundation (NSF) under Award No. 2153404. 
This work would not have been possible without the computing resources provided by the Ohio Supercomputer Center.
Thanks are also due to Qizhen Lan for providing feedback on the paper.

\bibliography{IEEEabrv,pruning}

\begin{thebibliography}{10}

\bibitem{gupta2015deep}
S.~Gupta, A.~Agrawal, K.~Gopalakrishnan, and P.~Narayanan, ``Deep learning with
  limited numerical precision,'' in {\em International conference on machine
  learning}, pp.~1737--1746, PMLR, 2015.

\bibitem{Hinton2015}
G.~E. Hinton, O.~Vinyals, and J.~Dean, ``Distilling the knowledge in a neural
  network,'' {\em CoRR}, vol.~abs/1503.02531, 2015.

\bibitem{Han2015}
S.~Han, J.~Pool, J.~Tran, and W.~J. Dally, ``Learning both weights and
  connections for efficient neural networks,'' in {\em Neural Information
  Processing Systems (NeurIPS)}, vol.~1, p.~1135–1143, 2015.

\bibitem{He2019}
Y.~He, P.~Liu, Z.~Wang, Z.~Hu, and Y.~Yang, ``Filter pruning via geometric
  median for deep convolutional neural networks acceleration,'' in {\em IEEE
  Conference on Computer Vision and Pattern Recognition (CVPR)},
  pp.~4335--4344, 2019.

\bibitem{Park2017}
J.~Park, S.~R. Li, W.~Wen, P.~T.~P. Tang, H.~H. Li, Y.~Chen, and P.~K. Dubey,
  ``Faster cnns with direct sparse convolutions and guided pruning,'' in {\em
  International Conference of Learning Representation (ICLR)}, 2017.

\bibitem{Han2016b}
S.~Han, X.~Liu, H.~Mao, J.~Pu, A.~Pedram, M.~Horowitz, and W.~J. Dally, ``Eie:
  Efficient inference engine on compressed deep neural network,'' {\em ACM/IEEE
  43rd Annual International Symposium on Computer Architecture (ISCA)},
  pp.~243--254, 2016.

\bibitem{Li2017}
H.~Li, A.~Kadav, I.~Durdanovic, H.~Samet, and H.~P. Graf, ``Pruning filters for
  efficient convnets,'' in {\em International Conference on Learning
  Representations (ICLR)}, 2017.

\bibitem{Molchanov2017}
P.~Molchanov, S.~Tyree, T.~Karras, T.~Aila, and J.~Kautz, ``Pruning
  convolutional neural networks for resource efficient inference,'' in {\em
  International Conference on Learning Representations (ICLR)}, 2017.

\bibitem{Tian2017deep}
Q.~Tian, T.~Arbel, and J.~J. Clark, ``Deep lda-pruned nets for efficient facial
  gender classification,'' in {\em Proceedings of the IEEE Conference on
  Computer Vision and Pattern Recognition Workshops}, pp.~10--19, 2017.

\bibitem{Tian2021task}
Q.~Tian, T.~Arbel, and J.~J. Clark, ``Task dependent deep lda pruning of neural
  networks,'' {\em Computer Vision and Image Understanding}, vol.~203,
  p.~103154, 2021.

\bibitem{LeCun1989}
Y.~LeCun, J.~S. Denker, and S.~A. Solla, ``Optimal brain damage,'' in {\em
  Neural Information Processing Systems (NIPS)}, p.~598–605, 1989.

\bibitem{Hassibi1992}
B.~Hassibi and D.~G. Stork, ``Second order derivatives for network pruning:
  Optimal brain surgeon,'' in {\em Neural Information Processing Systems
  (NIPS}, p.~164–171, 1992.

\bibitem{Srinivas2015}
S.~Srinivas and R.~V. Babu, ``Data-free parameter pruning for deep neural
  networks,'' in {\em British Machine Vision Conference (BMVC)}, 2015.

\bibitem{Chen2015}
W.~Chen, J.~T. Wilson, S.~Tyree, K.~Q. Weinberger, and Y.~Chen, ``Compressing
  neural networks with the hashing trick,'' in {\em International Conference on
  Machine Learning (ICML)}, p.~2285–2294, 2015.

\bibitem{Zhang2018}
T.~Zhang, S.~Ye, K.~Zhang, J.~Tang, W.~Wen, M.~Fardad, and Y.~Wang, ``A
  systematic dnn weight pruning framework using alternating direction method of
  multipliers,'' in {\em European Conference on Computer Vision (ECCV)},
  p.~184–199, 2018.

\bibitem{Hu2016}
H.~Hu, R.~Peng, Y.-W. Tai, and C.-K. Tang, ``Network trimming: A data-driven
  neuron pruning approach towards efficient deep architectures,'' {\em CoRR},
  vol.~abs/1607.03250, pp.~1--9, 2016.

\bibitem{Lin2020}
M.~Lin, R.~Ji, Y.~Wang, Y.~Zhang, B.~Zhang, Y.~Tian, and L.~Shao, ``Hrank:
  Filter pruning using high-rank feature map,'' in {\em IEEE Conference on
  Computer Vision and Pattern Recognition (CVPR)}, pp.~1526--1535, 2020.

\bibitem{He2018}
Y.~He, G.~Kang, X.~Dong, Y.~Fu, and Y.~Yang, ``Soft filter pruning for
  accelerating deep convolutional neural networks,'' in {\em International
  Joint Conference on Artificial Intelligence (IJCAI)}, p.~2234–2240, 2018.

\bibitem{Sui2021}
Y.~Sui, M.~Yin, Y.~Xie, H.~Phan, S.~Zonouz, and B.~Yuan, ``Chip: Channel
  independence-based pruning for compact neural networks,'' in {\em Neural
  Information Processing Systems (NeurIPS)}, 2021.

\bibitem{Xie2020}
Z.~Xie, L.~Zhu, L.~Zhao, B.~Tao, L.~Liu, and W.~Tao, ``Localization-aware
  channel pruning for object detection,'' {\em Neurocomputing}, vol.~403,
  pp.~400--408, 2020.

\bibitem{Li2022}
S.~Li, L.~Xue, L.~Feng, Y.~Wang, and D.~Wang, ``Object detection network
  pruning with multi-task information fusion,'' {\em World Wide Web}, vol.~25,
  p.~1667–1683, 2022.

\bibitem{Simonyan14}
K.~Simonyan, A.~Vedaldi, and A.~Zisserman, ``Deep inside convolutional
  networks: Visualising image classification models and saliency maps,'' in
  {\em Workshop at International Conference on Learning Representations
  (ICLRW)}, 2014.

\bibitem{Springenberg15}
J.~T. Springenberg, A.~Dosovitskiy, T.~Brox, and M.~A. Riedmiller, ``Striving
  for simplicity: The all convolutional net,'' in {\em Workshop at
  International Conference on Learning Representations (ICLRW)}, 2015.

\bibitem{Zhou2016}
B.~Zhou, A.~Khosla, A.~Lapedriza, A.~Oliva, and A.~Torralba, ``Learning deep
  features for discriminative localization,'' in {\em IEEE Conference on
  Computer Vision and Pattern Recognition (CVPR)}, (Los Alamitos, CA, USA),
  pp.~2921--2929, IEEE Computer Society, jun 2016.

\bibitem{Selvaraju2017}
R.~R. Selvaraju, M.~Cogswell, A.~Das, R.~Vedantam, D.~Parikh, and D.~Batra,
  ``Grad-cam: Visual explanations from deep networks via gradient-based
  localization,'' in {\em IEEE International Conference on Computer Vision
  (ICCV)}, pp.~618--626, 2017.

\bibitem{Chattopadhay2018}
A.~Chattopadhay, A.~Sarkar, P.~Howlader, and V.~N. Balasubramanian,
  ``Grad-cam++: Generalized gradient-based visual explanations for deep
  convolutional networks,'' in {\em IEEE Winter Conference on Applications of
  Computer Vision (WACV)}, pp.~839--847, 2018.

\bibitem{Ge2021}
Z.~Ge, S.~Liu, F.~Wang, Z.~Li, and J.~Sun, ``{YOLOX:} exceeding {YOLO} series
  in 2021,'' {\em CoRR}, vol.~abs/2107.08430, 2021.

\bibitem{Chen2021}
Q.~Chen, Y.~Wang, T.~Yang, X.~Zhang, J.~Cheng, and J.~Sun, ``You only look
  one-level feature,'' in {\em IEEE Conference on Computer Vision and Pattern
  Recognition (CVPR)}, pp.~13039--13048, 2021.

\bibitem{Redmon2013}
J.~Redmon, ``Darknet: Open source neural networks in c.''
  \url{http://pjreddie.com/darknet/}, 2013-2016.

\bibitem{He2016}
K.~He, X.~Zhang, S.~Ren, and J.~Sun, ``Deep residual learning for image
  recognition,'' in {\em IEEE Conference on Computer Vision and Pattern
  Recognition (CVPR)}, pp.~770--778, 2016.

\bibitem{Lin2014}
T.-Y. Lin, M.~Maire, S.~J. Belongie, J.~Hays, P.~Perona, D.~Ramanan,
  P.~Doll{\'a}r, and C.~L. Zitnick, ``Microsoft coco: Common objects in
  context,'' in {\em European Conference on Computer Vision}, 2014.

\bibitem{Geiger2012}
A.~Geiger, P.~Lenz, and R.~Urtasun, ``Are we ready for autonomous driving? the
  kitti vision benchmark suite,'' in {\em IEEE Conference on Computer Vision
  and Pattern Recognition (CVPR)}, pp.~3354--3361, 2012.

\bibitem{Qizhen2022}
Q.~Lan and Q.~Tian, ``Instance, scale, and teacher adaptive knowledge
  distillation for visual detection in autonomous driving,'' {\em IEEE
  Transactions on Intelligent Vehicles}, pp.~1--14, 2022.

\bibitem{Choi2022}
J.~I. Choi and Q.~Tian, ``Adversarial attack and defense of yolo detectors in
  autonomous driving scenarios,'' in {\em IEEE Intelligent Vehicles Symposium
  (IV)}, pp.~1011--1017, 2022.

\bibitem{Liu2017}
Z.~Liu, J.~Li, Z.~Shen, G.~Huang, S.~Yan, and C.~Zhang, ``Learning efficient
  convolutional networks through network slimming,'' in {\em IEEE international
  conference on computer vision (CVPR)}, pp.~2736--2744, 2017.

\end{thebibliography}

\end{document}